\pdfoutput=1

\documentclass[11pt]{article}

\usepackage{acl}

\usepackage{times}
\usepackage{latexsym}

\usepackage{graphicx}
\usepackage[shortlabels]{enumitem}
\usepackage{float}
\usepackage{multicol}
\usepackage{multirow}
\usepackage{listings}
\usepackage{color}
\usepackage{subcaption}

\usepackage[T1]{fontenc}

\usepackage[utf8]{inputenc}

\usepackage{microtype}

\definecolor{darkgreen}{rgb}{0.33, 0.42, 0.18}
\lstdefinelanguage{LispSchemas}{
  language     = Lisp,
  keywords={header, types, rigid-conds, static-conds, pre-conds, post-conds, goals, episodes, necessities, certainties},
}
\lstdefinelanguage{DialogueExamples}{
    language = Lisp,
    keywords={},
    otherkeywords={SOPHIE, DAVID},
    keywords = [2]{User},
    keywordstyle=\color{purple},
    keywordstyle=[2]\color{blue},
}

\title{A Flexible Schema-Guided Dialogue Management Framework: From Friendly Peer to Virtual Standardized Cancer Patient}

\author{
  Benjamin Kane \and Catherine Giugno \and Lenhart Schubert \\
  University of Rochester \\
  \texttt{\{bkane2@ur, cgiugno@u, schubert@cs\}.rochester.edu} \\\AND
  Kurtis Haut \and Caleb Wohn \and Ehsan Hoque \\
  University of Rochester \\
  \texttt{\{khaut@cs, cwohn@u, mehoque@cs\}.rochester.edu}
}

\begin{document}
\maketitle
\begin{abstract}
A schema-guided approach to dialogue management has been shown in recent work to be effective in creating robust customizable virtual agents capable of acting as friendly peers or task assistants. However, successful applications of these methods in open-ended, mixed-initiative domains remain elusive -- particularly within medical domains such as virtual standardized patients, where such complex interactions are commonplace -- and require more extensive and flexible dialogue management capabilities than previous systems provide. In this paper, we describe a general-purpose schema-guided dialogue management framework used to develop SOPHIE, a virtual standardized cancer patient that allows a doctor to conveniently practice for interactions with patients. We conduct a crowdsourced evaluation of conversations between medical students and SOPHIE. Our agent is judged to produce responses that are natural, emotionally appropriate, and consistent with her role as a cancer patient. Furthermore, it significantly outperforms an end-to-end neural model fine-tuned on a human standardized patient corpus, attesting to the advantages of a schema-guided approach.
\end{abstract}

\section{Introduction}
\label{sec:introduction}


The schema-guided approach to dialogue management aims to provide a middle ground between end-to-end neural dialogue models and classical plan-based architectures; it allows for flexible management of dialogue, through matching of observed utterances to explicitly specified dialogue policies or ``schemas''. Recent work has demonstrated that a schema-guided approach is effective in task-specific dialogue systems \citep{rastogi-2020-schema-guided, mosig-2020-star, mehri-2021-schema-guided} as well as friendly peer systems capable of casual conversation across many topics \citep{razavi-2019-lissa}.

However, these previous systems were limited to relatively simple domains in which conversations are assumed to be single-initiative, and conversations are either shallow or task-directed in scope. Yet, many potential applications of dialogue technology are in domains where dialogue is complex and \textit{mixed-initiative} -- i.e., the user and system may both take the initiative at various times. These types of dialogues have radically different structure than task-oriented dialogues, requiring a greater degree of collaborative planning between the two agents \cite{walker-1990-mixed-initiative}. Furthermore, more complex domains may require a greater degree of semantic understanding by a dialogue system.

One particular domain in which successful applications of dialogue technology remain sparse is that of conversational virtual standardized patients (VSPs) that simulate realistic doctor-patient conversations for use in training or evaluating medical professionals. Despite the challenging nature of this domain, such systems could be broadly impactful: Communication skills on the part of the doctor are a well-recognized determinant of patient satisfaction and optimal outcomes \cite{korsch-1972-doctor, ha-2010-doctor, riedl-2017-influence, stewart-1995-effective, begum-2014-doctor}; yet providing communication training for doctors is a resource-intensive process, requiring extensive training for the role as well as taxing emotional investment from patient actors. As a result, the quality and availability of communication training are often quite variable. The ability to create, customize, and deploy virtual conversational patients that are capable of having realistic conversations with doctors could greatly alleviate these costs.

In this paper, we present a novel dialogue architecture used to create SOPHIE (Standardized Online Patient for Health Interaction Education), a system that allows doctors to hold conversations with a virtual cancer patient while receiving automated feedback on their communication skills. Our contributions include a dialogue manager based on \textit{dialogue schemas} -- explicit representations of \textit{expected} sequences of dialogue acts -- and a unique method of ``gist clause'' extraction as a preliminary step in mapping user utterances to semantic representations. The dialogue capabilities enabled by our approach were tested in a pilot study of the SOPHIE VSP. A crowdsourced evaluation of 30 resulting dialogue transcripts showed SOPHIE's performance to be superior to that of a large language model fine-tuned on a corpus of dialogues with human standardized patients.

\section{Related Work}
\label{sec:related_work}

The design of virtual humans for therapeutic use and conversational coaching has become a very active research area. One of the earliest such systems, the ``How Was Your Day'' avatar, used a template-based approach to generate empathetic responses in a conversation about the user's workday \citep{pulman-2010-howwasyourday}. Along these lines, the ERICA android \citep{ishii-2021-erica} was created to hold empathetic dialogues to help users at risk of isolation, although this work focused largely on non-verbal aspects of communication. The SimSensei avatar \citep{devault-2014-simsensei} uses a classifier-based approach to provide healthcare support to users, while the INOTS avatar \citep{campbell-2011-inots} uses a similar approach to coach naval officers on interpersonal skills. The LISSA system \citep{razavi-2016-lissa, razavi-2019-lissa} allows users to practice social skills through casual conversation with a virtual avatar, using a schema-guided framework.

More recently, end-to-end neural language models have been effective in creating chatbots for entertainment purposes \citep{adiwardana-2020-meena, roller-2020-blenderbot}, though the ``black box'' nature of these models has constrained their use in sensitive domains. Contemporary work on schema-guided dialogue has shown that robust zero-shot dialogue behavior could be achieved by fine-tuning language models to fill in schemas consisting of slot-value pairs \citep{rastogi-2020-schema-guided} or to align schemas consisting of conversation graphs to observed user utterances \citep{mosig-2020-star, mehri-2021-schema-guided}. In contrast with this work, we employ a schema-guided approach in an open-ended virtual standardized patient domain as opposed to task-specific dialogues, and rely on a semantically rich event schema representation instead of slot-value pairs or conversation graphs.

Previous work on virtual standardized patients has demonstrated that even relatively simple avatars, such as the early DIANA system, can offer comparable training benefits to human standardized patients \citep{lok-2006-diana}. Similarly, the SIDNIE avatar was created to train nursing students to conduct interviews with young patients, and was effective in improving its users' skills despite using menu-based interaction as opposed to natural dialogue \citep{dukes-2013-sidnie}. Other virtual patient research has focused on enabling more natural dialogue behavior for VSPs across several different domains, typically through the use of pattern-matching and classifier-based methods \citep{rossen-2009-human-centered, carnell-2015-adapting, maicher-2017-standardized-patient, rakofsky-2020-virtual}. However, these systems are generally designed around single-initiative domains, such as history-taking conversations (in which the doctor has initiative and needs to solicit information from the patient); to our knowledge, there is no VSP system that can support open-ended mixed-initiative dialogue such as that of a cancer patient discussing their future with a doctor.

\section{Dialogue Manager Overview}
\label{sec:dialogue_manager}

\begin{figure*}[ht]
    \centering
    \includegraphics[width=0.8\textwidth]{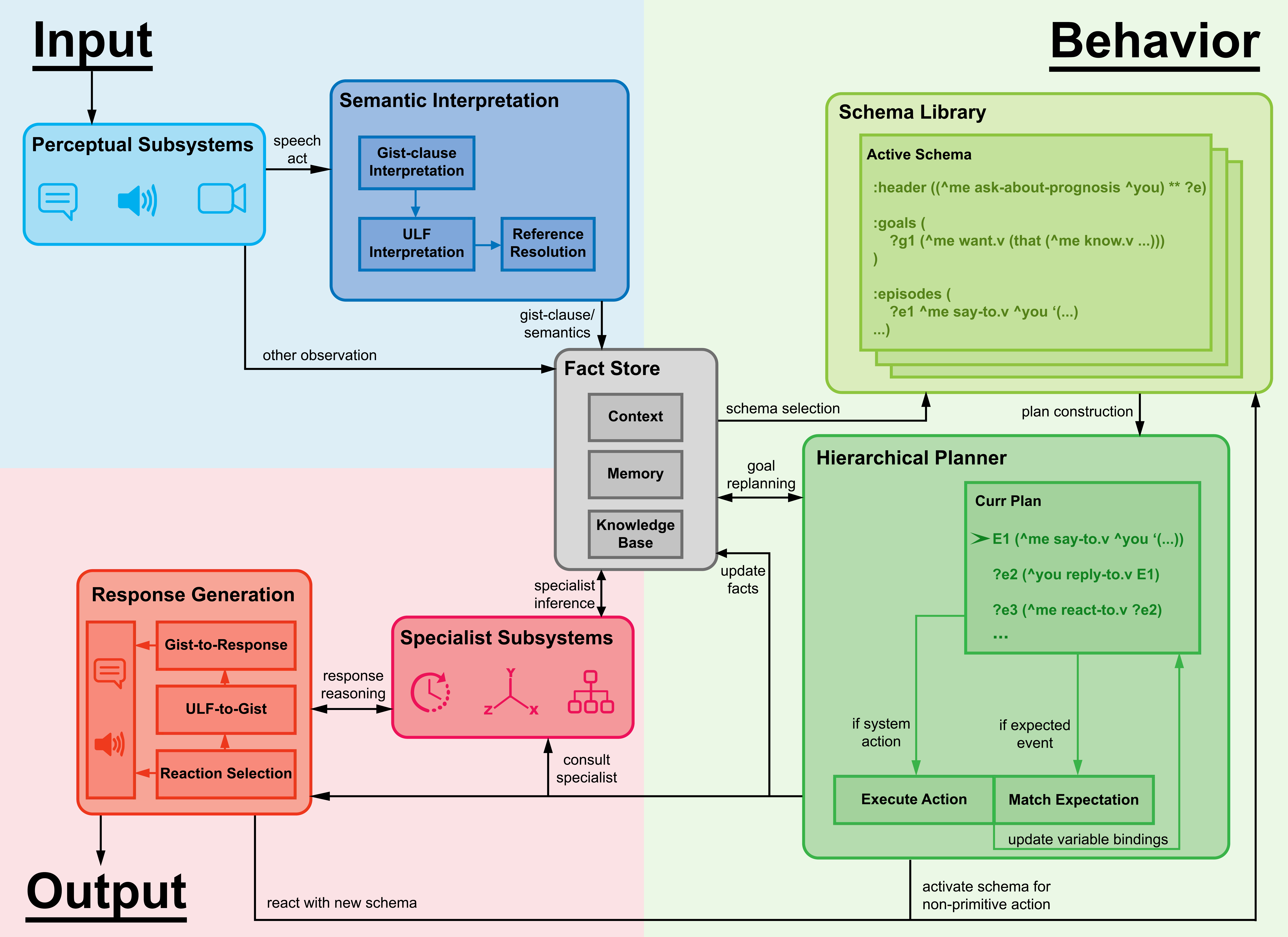}
    \caption{Eta system architecture.}
    \label{fig:architecture}
\end{figure*}

In this section, we provide an overview of the Eta dialogue manager: a general-purpose, modular, schema-based dialogue management framework, diagrammed in Figure \ref{fig:architecture}. Similarly to the LISSA system \citep{razavi-2017-managing-dialogue-schemas}, the Eta dialogue manager relies centrally on a library of \textit{dialogue schemas} -- representations of prototypical dialogues -- to flexibly plan dialogues, as well as \textit{transduction trees} for interpretation and response generation. In contrast to the LISSA system, Eta is a general-purpose dialogue shell that is capable of alternating between multiple parallel tasks including observation, interpretation, inference, and planning. In order to create an avatar using Eta, a designer need only create a set of dialogue schemas and transduction trees; many aspects of these trees are portable across domains. We discuss the design of the SOPHIE avatar in Section \ref{sec:evaluation}.

\subsection{Dialogue Schemas}
\label{subsec:schemas}

The concept of ``schemas'' has a long history in AI research, arising from attempts to capture generic and prototypical knowledge in a declarative symbolic form \citep{van-dijk-kintsch-1983-strategies,minsky-1974-knowledge,schank-abelson-1977-scripts}. A schema is typically conceptualized as a packet of \textit{expected} properties of an object or event, such that a strong match between an observed object and properties of a schema allows one to make inferences about other properties associated with that schema. Event schemas also represent a generalization of hierarchical plans in the sense that they can be used for classic planning through goal chaining and step decomposition, but also more ``reflexive'' planning based on expectation-matching and abductive inference \citep{turner-1994-schema}.

The Eta dialogue manager relies on a library of \textit{dialogue schemas}: event schemas for various dialogues that an avatar might encounter. Formally, our schema representation is based on Episodic Logic (EL) -- a type-coherent logical semantic representation that closely matches the form and expressive capacity of natural English. A dialogue schema minimally consists of a series of expected episodes (assumed to occur sequentially by default, though other temporal orderings are possible), but may also contain other schema sections containing EL formulas for other expected properties of the dialogue -- participant roles, preconditions, goals, etc\footnote{A full list of supported schema sections is shown in Appendix \ref{app:schemas}.}. Episodes within a dialogue schema may correspond to expected user inputs, observations, primitive actions that the avatar is capable of executing, or a sub-schema to activate.

An example of a dialogue schema from the SOPHIE avatar (in which dialogue schemas are used to represent topical ``sub-dialogues'') is shown in Figure \ref{fig:schema-example}. Symbols beginning with `?' represent variables that are replaced with constants upon instantiation of a schema, while `\^{}me' and `\^{}you' indexically refer to the dialogue participants. The schema begins with a header that provides a characterizing formula for the head episode `?e', and is followed by a schema section specifying goals for that episode (viz., that the system learns the answer to the quoted question). The body of the schema specifies that the sub-dialogue consists of a sequence of actions: The avatar asks the user some variant of the quoted question; the user replies to the avatar, and the avatar reacts to the user's reply.


\begin{figure*}[t]
\small
\begin{lstlisting}[language=LispSchemas,frame=single,mathescape=true]
(event-schema :header ((^me ask-about-test-results.v ^you) ** ?e)
;`````````````````````````````````````````````````````````````````````````
:goals (
  ?g1 (^me want.v (that (^me know.v (ans-to '(What do my test results mean ?))))))

:episodes (
  ?e1 (^me paraphrase-to.v ^you '(What do my test results mean ?))
  ?e2 (^you reply-to.v ?e1)
  ?e3 (^me react-to.v ?e2)))
\end{lstlisting}
\caption{An example of a dialogue schema for SOPHIE asking about her test results.}
\label{fig:schema-example}
\end{figure*}

\subsection{Fact Store}

The various components of Eta are built around a central \textit{fact store}. This component is divided into a dialogue context (containing facts in the common ground, assumed to be ``true now''), an episodic memory, and a knowledge base for general facts. Each contains EL formulas, indexed on their predicates and arguments for efficient retrieval.

\subsection{Perceptual and Specialist Subsystems}

Eta supports integration with various domain-specific \textit{perceptual} and \textit{specialist} subsystems. Perceptual subsystems, which include text interfaces, speech recognition programs, and vision systems, are used to collect observations, some of which are interpreted by the semantic interpretation module. Specialist subsystems, which include temporal models, spatial models, and type ontologies, may be used as a subroutine in inference or response generation. Both types of subsystems interact with the core Eta dialogue manager through EL queries.

\subsection{Semantic Interpretation}
\label{subsec:interpretation}

After observing an utterance from the user, Eta attempts to interpret its semantic meaning using the context of the immediately preceding speech act. Semantic interpretation is performed at two levels, allowing for both shallow and deep understanding. First, an explicit context-independent \textit{gist-clause} is extracted from the user's input representing the \textit{gist} of the user's utterance. The intuition behind this stage of interpretation is that the context of the immediately preceding turn is typically sufficient to resolve many context-sensitive phenomena (e.g., ellipsis, simple anaphora, ambiguity, etc.) using relatively superficial text processing; mapping a complex utterance to a context-independent `canonical' form greatly simplifies semantic parsing and response generation. In the case of the SOPHIE avatar, this shallow stage of interpretation is often sufficient for response generation.

Optionally, Eta can then parse a gist-clause into an Unscoped Logical Form (ULF) representation -- a variant of EL that leaves tense operators and quantifiers unscoped to simplify the representation. In this case, a constraint-based reference resolution module attempts to resolve any references not handled during the gist-clause interpretation.

In practice, gist-clause interpretation is done using the \textit{hierarchical pattern transduction tree} method previously introduced by the LISSA system \citep{razavi-2017-managing-dialogue-schemas}. Transduction trees specify patterns at their nodes, with branches from a node providing alternative continuations as a hierarchical match proceeds. Terminal nodes have associated \textit{directives} indicating whether they provide a gist-clause template, send input to some subordinate tree, or some other outcome. The pattern nodes use simple template-like patterns that look for particular words or word features, and allow for ``match-anything'', length-bounded word spans.




ULF interpretation can likewise be performed using hierarchical pattern transduction tree methods through the use of phrase-based trees with recursive compositional directives at leaf nodes. This provides a reliable but restricted method for semantic parsing; future efforts to integrate a generic semantic parser are discussed in Section \ref{sec:limitations_future}.

\subsection{Hierarchical Planner}

The hierarchical planner maintains a pointer to the current dialogue plan, represented as a graph of plan steps, each consisting of a pointer to the previous step, next step, and a subplan (when applicable). Plans are dynamically constructed and modified over the course of the dialogue depending on the conditions of the respective schema. For instance, if a schema contains a goal condition that is already satisfied, the system may skip over it. On the other hand, if the end of the schema is reached and the goal still isn't satisfied, the system can attempt to replan according to some fallback strategy\footnote{Currently, these backup strategies are determined using pattern transduction trees, and may consist of generating a response, activating a new sub-schema, or simply moving on.}.

If the currently pending step in the plan is an expected user action, the system attempts to \textit{match that expectation} to a ``currently true'' fact in context, unifying any variables in the corresponding formula. If no fact is matched, the plan will not proceed unless some elapsed time since the last plan modification is exceeded, determined by the certainty of the expected episode in the schema.

If the currently pending step is a system action, the system will \textit{execute the action}. If the action is non-primitive, it will select the appropriate sub-schema and expand it as a subplan to the current step, or otherwise modify the plan directly. If the action is primitive, it can have any of a number of effects depending on the implementation of the function for that action: It might update the fact store; it might request domain-specific reasoning from a specialist subsystem; it might request a perceptual subsystem to attend to a specific entity; or it might trigger response generation.

\subsection{Response Generation}
\label{subsec:generation}

The system's response generation is governed by a hierarchical pattern transduction tree for selecting the system's reaction to the user's previous gist-clause. As in the case of interpretation, the terminal nodes of transduction trees specify directives which may include directly outputting some template, or selecting a sub-schema to activate and react with.

In more complex cases where the system itself forms a response gist-clause (this may either be generated from a response ULF, or part of a ``paraphrase'' action in a schema, such as in Figure \ref{fig:schema-example}), the system mirrors the process used for gist-clause interpretation: The user's previous gist-clause will be used as context for paraphrasing the system's gist-clause into a more natural form.


\section{SOPHIE Evaluation}
\label{sec:evaluation}
\subsection{Design of SOPHIE Avatar}


The SOPHIE avatar\footnote{The SOPHIE avatar was created using the SitePal API: https://www.sitepal.com.} was designed to represent an elderly lung cancer patient who had had previous testing done, and is meeting with a doctor to learn more about her test results and prognosis. Users interact with the system verbally, with a manually calibrated silence threshold determining turn-taking, and an off-the-shelf automatic speech recognizer (ASR) converting audio to text input.

The dialogue capabilities of the avatar were enabled through the creation of dialogue schemas for various topics that the avatar may discuss (such as the example in Figure \ref{fig:schema-example}), a top-level dialogue schema containing the expected sequence of topics that the avatar will ask about in a session (flexible to modification during the conversation), and a set of pattern transduction trees for interpretation and response generation. These components were hand-engineered in consultation with palliative care experts; in order to anticipate possible inputs and create realistic responses, we also used a corpus of human SP conversation transcripts from the VOICE dataset \citep{hoerger-2013-values}, as well as feedback and transcripts from a small focus group study involving medical students interacting with SOPHIE \citep{ali-2021-sophie}.

Our pattern transduction-based approach allows for easy customization by dialogue designers who are not AI experts. A non-expert was able to learn the pattern-matching language and independently create new transduction trees within a week, and on average were able to create a transduction tree for a new topic in approximately 15-30 minutes (depending on the level of detail required and whether or not the new topic is related to any previous topics).

\subsection{Evaluation Method}

In the following sections, we describe our evaluation of the SOPHIE avatar using both expert and crowdsourced annotations of turns from SOPHIE conversations, including a comparison to responses generated by a neural baseline in the same domain.

\subsection{Pilot Dataset}


We conducted an initial pilot experiment with SOPHIE consisting of 30 medical student participants, half of which were randomly assigned to a treatment group who interacted with SOPHIE for two sessions each; the other half were assigned to a control group who had no interaction with the system\footnote{The pilot experiment was intended to test the effects of the SOPHIE feedback system on user communication skills, though this lies outside the scope of our paper.}. We obtained a dataset consisting of 397 turns across 30 conversation sessions from 15 participants.



\subsubsection{Expert Annotations}

As a preliminary form of evaluation, the researchers involved in the design of the pattern transduction trees for SOPHIE independently provided expert turn-by-turn annotations on whether the system extracted a correct gist-clause, an incorrect gist-clause, or failed to extract a gist-clause; as well as whether the system gave an appropriate response, inappropriate response, or a non-contentful clarification request. They also annotated any significant ASR errors that were observed, which fell into two categories: transcription errors and turn-taking errors (i.e., where the user was cut off).


\subsubsection{Neural Baseline}


We also establish a ``neural baseline'' for our conversation domain -- that is, the performance of a state-of-the-art language model fine-tuned on human SP transcripts -- and compare the responses generated by this model, when prompted with turns from the pilot dataset, to the responses from our system. Specifically, we fine-tuned DialoGPT-medium \citep{zhang-2020-dialogpt} on the VOICE human SP dataset \citep{hoerger-2013-values} -- filtering for only patient turns, with a context window size of 5 previous utterances -- for 5 epochs, using a batch size of 1. This resulted in a validation set perplexity of 6.53; gains after 5 epochs were negligible. To generate model responses for the pilot data, for each turn in the dataset, we concatenated the user utterance with the context of the immediately preceding utterance (separated by end-of-turn tokens) and let the model generate the next response. We used a length penalty of 0.5 and a repetition penalty of 1.4 as generation parameters; these were found to generate the best responses through manual inspection.

%
%


\subsubsection{Crowdsourced Evaluation}

Given the SOPHIE responses and corresponding responses generated by the neural baseline model, we crowdsourced annotations on response quality using Amazon Mechanical Turk. We first removed 89 turns that were judged by either of the expert annotators to include significant ASR errors in the doctor's input. The remaining turns were used to form 308 items, each consisting of the context of the previous patient utterance, the doctor's utterance, the response generated by Eta, and the response generated by the neural baseline model.

Items were randomly distributed into 20 Human Intelligence Tasks (HITs), each containing 16 items. To avoid introducing possible annotation bias, items were exactly balanced on string length of each text field (i.e., items were assigned to four high/low bins for each text field computed using a median split), and were approximately balanced on the expert annotations of response quality (ensuring that each HIT has about as many good responses as bad responses from each system). For each item, workers were shown a response from Patient A and the previous two dialogue turns, and then asked to rate the following four questions about the response on a Likert scale. Then workers were shown a response from Patient B to the same previous turns, and asked to rate the same four questions. The Eta responses and neural baseline responses were randomly assigned to Patient A and Patient B.

\begin{enumerate}[noitemsep,label=Q\arabic*.]
    \item Patient A/B's response is fluent and natural.
    \item Patient A/B's response is consistent with her having understood the preceding turns.
    \item Patient A/B's response is consistent with her role as a cancer patient.
    \item Patient A/B's response expresses appropriate emotions (if an emotional response).
\end{enumerate}

Workers were instructed to focus on the quality of the response irrespective of the patient background and previous turns for answering Q1, and to overlook fluency in the response for answering Q2-Q4. Additionally, Q3-4 had ``Not Applicable'' options, in the event that these questions could not be evaluated (for instance, Q3 is impossible to evaluate if the patient asks \textit{``Could you repeat that?''}, and Q4 is impossible to evaluate if the patient says \textit{``I'm taking Lortab for the pain.''}); any Not Applicable responses were discarded.

\subsection{Results}

\begin{figure*}
    \centering
    \includegraphics[width=\textwidth]{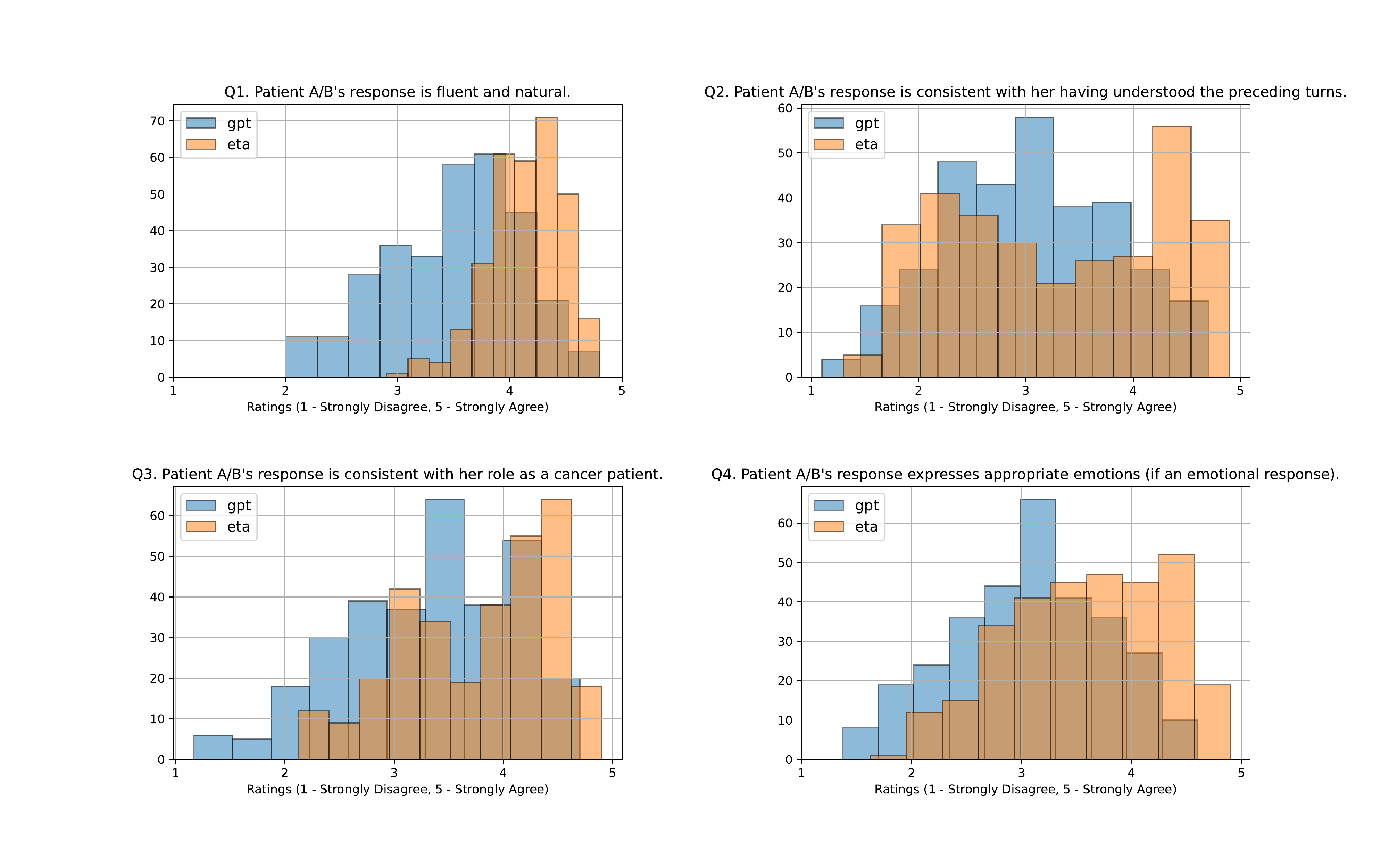}
    \caption{Distributions of average crowdsourced response ratings for each question.}
    \label{fig:sophie-mturk-results}
\end{figure*}





Table \ref{tab:sophie-expert-results} shows the averaged expert annotations (N=2) for ASR errors, the fraction of correctly extracted gist-clauses, and the fraction of appropriate responses from the system (expected to be higher than the former quantity because the system might fail to extract a gist-clause but still give an appropriate ``default'' response). These statistics are corrected for ASR errors by assuming that a clarification request by the system is the correct behavior when the input contains a significant ASR error and no gist clause is extracted. We also show the fraction of responses judged as appropriate, conditioned on a gist-clause having been extracted. Interannotator agreement (Cohen's kappa) was quite high for both annotations, at 0.85 for gist-clause annotations and 0.71 for response annotations.

From the crowdsourced Amazon Turk study, we obtained 10 ratings for questions Q1-4 for each item, from 167 native English speakers. The resulting distributions of ratings per item (averaged across participants, discarding `N/A' values) are shown as histograms in Figure \ref{fig:sophie-mturk-results} (a score of 1 corresponds to ``Strongly Disagree'', and 5 to ``Strongly Agree''). The mean and median scores for each question across all items are shown in Table \ref{tab:sophie-mturk-results}; the mean differences between Eta and DialoGPT were found to be statistically significant for each question (P < 0.05) using a Mann–Whitney U test.

\begin{table}[h!]
\centering
\begin{tabular}{| p{5cm} | c |} 
    \hline
    ASR Errors & 20\% \\
    \hline\hline
    Correct Gist Extracted & 39\% \\
    \hline
    No Gist Extracted & 41\% \\
    \hline
    Incorrect Gist Extracted & 20\% \\
    \hline\hline
    Appropriate Response & 49\% \\
    \hline
    Clarification Request Response & 28\% \\
    \hline
    Inappropriate Response & 24\% \\
    \hline\hline
    Appropriate Response Given Gist Clause Extracted & 72 \% \\
    \hline
\end{tabular}
\caption{Results from the preliminary expert annotations.}
\label{tab:sophie-expert-results}
\end{table}

\subsection{Discussion}

From the expert annotations in Table \ref{tab:sophie-expert-results}, we observe that the rates of correct gist-clause extraction and appropriate response generation were 39\% and 49\% respectively. In many cases, failure to extract a correct gist-clause would also lead to a non-contentful clarification request rather than an appropriate response. However, we observe that in the cases where the system was able to extract a gist-clause, the fraction of appropriate responses by the system was significantly higher -- suggesting that the primary bottleneck in the current SOPHIE system is the gist-clause extraction step.

\begin{table}[h!]
\centering
\begin{tabular}{| c | c | c | c | c | c |} 
    \hline
    \multicolumn{2}{|c|}{Agent}    & Q1 & Q2 & Q3 & Q4 \\
    \hline
    \multirow{2}{*}{GPT}  & Mean   & 3.49 & 3.03 & 3.30 & 3.06 \\
                          & Median & 3.55 & 3.00 & 3.40 & 3.11 \\
    \hline
    \multirow{2}{*}{Eta}  & Mean   & 4.15 & 3.29 & 3.78 & 3.60 \\
                          & Median & 4.20 & 3.20 & 4.00 & 3.60 \\
    \hline
    \multirow{2}{*}{Diff} & Mean   & +0.66 & +0.26 & +0.48 & +0.54 \\
                          & Median & +0.65 & +0.20 & +0.60 & +0.49 \\
    \hline
    
\end{tabular}
\caption{Mean and median of average crowdsourced responses across items for each set of responses.}
\label{tab:sophie-mturk-results}
\end{table}

While the expert-annotated response accuracy is fairly low compared to results from previous VSP evaluations, which report between 60\% and 80\% response accuracy \citep{maicher-2017-standardized-patient}, we argue that our results are consistent with the greater overall complexity of our conversation domain compared to past systems, as discussed in Section \ref{sec:related_work}. Thus, we view our results as a useful baseline for future work in this domain.

From the crowdsourced ratings summarized in Table \ref{tab:sophie-mturk-results}, we observe that the SOPHIE avatar was rated higher on average for each question than the fine-tuned DialoGPT model, with the largest gains being in fluency/naturalness, consistency with the role of a patient, and appropriateness of emotions. We take these results to be an indication that the customizability of Eta provides a distinct advantage over an end-to-end deep learning approach, since our system allows the interpretive rules and responses to be designed with a particular role and character in mind. In contrast, the fine-tuned DialoGPT model would sometimes confuse its role in the dialogue (e.g., act as a doctor rather than a patient), an issue related to the tendencies of end-to-end neural dialogue models to hallucinate false or contradictory knowledge \citep{roller-2020-blenderbot}.

The lowest ratings for both sets of responses -- and the smallest gains by Eta above the neural baseline -- were concerning whether the agent demonstrated understanding of the previous turns. This illustrates that the superficial pattern-matching rules used by Eta are not, at least in their current form, sufficient to demonstrate understanding. Interestingly, the distribution of ratings for Eta on this question in Figure \ref{fig:sophie-mturk-results} is strongly bimodal, unlike DialoGPT. We believe that this pattern attests to the results found in the expert annotations, namely, that the \textit{conditional} probability of Eta generating a good response given that it extracted a gist-clause is high, but when the system fails to extract a gist-clause, it generally makes clarification requests or other inappropriate responses. The average rating for this question, then, could likely be improved by increasing the coverage of gist-clause extraction.


\section{Limitations and Future Work}
\label{sec:limitations_future}
As discussed in Section \ref{sec:evaluation}, our system is limited in its ability to successfully extract gist-clauses from the user's utterance, with failures frequently occurring due to insufficient pattern transduction trees, out-of-domain inputs, and minor ASR errors.

To account for scenarios where the system fails to extract a gist-clause, we employed two types of fallback strategies for response generation in the SOPHIE avatar: The system may either make a generic clarification request, or the system may give a default response (dependent on the particular schema) that gives a generic reaction to the user and pushes the conversation forward. Examples from the SOPHIE pilot for each of these two strategies are shown in Appendix \ref{app:dialogue-examples}. As can be seen in these examples, both fallback strategies have advantages in some scenarios, but fail in other scenarios. For instance, when users were asked to clarify an utterance, they would sometimes simply repeat the language in their previous utterance, leading to another gist-clause extraction failure. Moreover, a clarification request after a lengthy response by the user would cause frustration for the user. On the other hand, giving a generic default reaction may be inappropriate if the user asked a question, mentioned important information, or expressed empathy. Thus, neither strategy for response generation is sufficient for accounting for errors in the interpretation pipeline.

Finally, we note a limitation of our evaluation method: The annotations collected only assess \textit{local coherence}, i.e., whether a response is appropriate given the immediately preceding turn, and don't measure \textit{global coherence} across the whole dialogue. Since the current SOPHIE avatar relies largely on shallow pattern transduction, some issues with global coherence were also observed including a tendency to repeat certain responses, or to occasionally give inconsistent responses.

In the future, we aim to explore other fallback methods for generation -- such as detecting named entities or broad thematic topics from user input and responding to those using either templates or generative language models. Furthermore, we aim to make gist-clause extraction less brittle by experimenting with a constrained language model approach, as was found to be effective in other semantic parsing work \citep{shin-2021-constrained}. Finally, we aim to enable deeper semantic understanding through the incorporation of a neural English-to-ULF parser \cite{kim-2021-transition}, allowing for avatars like SOPHIE to draw on logical inferences to improve global coherence and activate relevant schemas.

\section{Conclusion}
In this paper, we described a flexible and general-purpose schema-guided dialogue framework used to create SOPHIE, a virtual standardized cancer patient that allows doctors to practice conversations with patients. We demonstrated the advantages of our dialogue framework through a crowdsourced evaluation of transcripts obtained from a pilot experiment with SOPHIE. We found that SOPHIE was judged to be fluent, natural, consistent with her patient role, and appropriate in her emotions; our system also outperformed a neural baseline model fine-tuned on a corpus of standardized cancer patient dialogues. However, both the crowdsourced evaluation and an expert annotation of gist-clause and response accuracy indicated that the system is significantly limited in its ability to understand the user. Future work will aim to explore methods to improve coverage and robustness of our semantic interpretation component, as well as better fallback strategies for response generation.

\bibliography{anthology,custom}
\bibliographystyle{acl_natbib}

\clearpage
\appendix

\section{Supported Schema Sections}
\label{app:schemas}
The schema sections defined by our dialogue schema representation are as follows\footnote{The section names are slightly modified from, but mostly analogous to, those discussed in \citep{lawley-2019-schema-story-understanding}}.

\begin{itemize}
    \item \textbf{episodes}: the minimal requirement for a dialogue schema is a list of episode variables and associated formulas. Episodes typically reflect speech acts by participants, but do not necessarily do so -- they could in principle be any anticipated event, a proposition expected to become true, or more complex procedural behavior such as repeating an episode until a contextual condition is met.
    \item \textbf{episode-relations}: temporal relations between episode variables specified in the episodes section. The default ordering between episodes is sequential in the order they occur in the schema, but other constraints can be specified (such as ``consec'' for two directly consecutive episodes, or ``same-time'' for simultaneous episodes).
    \item \textbf{types}: non-fluent\footnote{The terms ``fluent'' and ``non-fluent'' are borrowed from situation calculus to refer to predicates whose truth values are subject to change over time and those that remain fixed within the scope of a schema, respectively.} type predications for individuals occurring in the schema, e.g., that the user is a human.
    \item \textbf{rigid-conds}: any non-fluent predications about individuals occurring in the schema apart from the types, e.g., that the user is a doctor.
    \item \textbf{static-conds}: fluent predications which are expected to hold throughout the schema episode, e.g., that the avatar has cancer.
    \item \textbf{pre-conds}: fluent predications that are expected/required to be true at the initiation of the episode represented by the schema.
    \item \textbf{post-conds}: fluent predications that are expected/required to be true at the conclusion of the episode represented by the schema.
    \item \textbf{trigger-conds}: fluent predications which, if true, signal that the schema should be instantiated or at least marked as a candidate for instantiation. These likely overlap with other fluent schema conditions.
    \item \textbf{goals}: formulas corresponding to the goals of participants in the schema.
    \item \textbf{necessities}: associate values in [0,1] with schema conditions indicating how necessary it is that those conditions hold. For instance, if some precondition is necessary to degree 1, then it is strictly required to hold for the schema to be instantiated. If a precondition is necessary to degree 0.5 and doesn't hold, the schema might be dispreferred but could still be instantiated if other conditions are sufficiently compelling.
    \item \textbf{certainties}: associate values in [0,1] with schema episodes indicating how certain it is that those episodes will be observed. If an episode is certain to degree 1, then that episode must be matched to an observation for the schema to proceed. If an episode is certain to degree 0.5, the agent might try for a while to match the expectation to an observation, but otherwise may move on with the rest of the schema.
\end{itemize}

\section{SOPHIE Dialogue Examples}
\label{app:dialogue-examples}
Some ``cherry-picked'' dialogue excerpts from the SOPHIE pilot study are shown in Figure \ref{fig:sophie_good_transcripts}. Some ``lemon-picked'' examples of dialogues from the SOPHIE pilot experiment that show limitations of our current system, as discussed in Section \ref{sec:limitations_future}, are shown in Figure \ref{fig:sophie_bad_transcripts}. Minor ASR errors have been corrected in these examples for readability.

\begin{figure*}[t]
    \begin{subfigure}{\textwidth}
        \small
\begin{lstlisting}[language=DialogueExamples,frame=single,mathescape=true]
SOPHIE: I've already scanned the results, but don't understand what
        they are telling me. Could you explain what they mean?
User:   So unfortunately Sophie I have some bad news. It looks like the
        cancer has grown and spread.
SOPHIE: Those are not the words I wanted to hear. I mean, I was bracing
        for the worst, since I could tell by the pain that it's bad.
        But to hear that the cancer has spread is quite depressing.
        What does it all mean for me?
User:   How much information would you like to know about the prognosis?
SOPHIE: I feel very anxious about my condition. I feel like it's gotten
        worse. I want to know what this means for me.
User:   What concerns do you have about the future?
SOPHIE: What worries me the most is just how uncertain my future is.
        Some days I believe my condition will get better and others
        I brace myself for the worst. Is there any way you can let me
        know what I should expect?
...

\end{lstlisting}
        \caption{A ``cherry-picked'' excerpt from the SOPHIE pilot study in which the doctor informs SOPHIE that her cancer has spread, and discusses her concerns for her future.}
        \label{fig:sophie_good_transcript1}
     \end{subfigure}
     \begin{subfigure}{\textwidth}
        \small
\begin{lstlisting}[language=DialogueExamples,frame=single,mathescape=true]
SOPHIE: I still don't understand how this happened! Just three months
        ago I didn't even know I had cancer.
User:   I agree that lung cancer in particular is a very difficult
        diagnosis for patients and their families.
SOPHIE: Cancer is such a challenging disease to live with. But I
        appreciate all you've been doing to help me.
SOPHIE: I haven't told my family everything yet. I wanted to wait to
        talk to you first. What should I say to them?
User:   How much information do you want them to know?
SOPHIE: My family knows about my cancer already, but they don't
        really know how bad it is, or what it means for me. How should
        I discuss these with them?
User:   Being open and honest is going to help you the most, and if
        you'd like they can come to a visit with you and we can help
        you talk to them.
SOPHIE: Okay. It will be hard to give them such bleak news, but they
        deserve to know the truth.
...
\end{lstlisting}
        \caption{A ``cherry-picked'' excerpt from the SOPHIE pilot study in which the doctor gives SOPHIE tips on how to break the news to her family.}
        \label{fig:sophie_good_transcript2}
     \end{subfigure}
     \caption{}
     \label{fig:sophie_good_transcripts}
\end{figure*}



\begin{figure*}[t]
    \begin{subfigure}{\textwidth}
        \small
\begin{lstlisting}[language=DialogueExamples,frame=single,mathescape=true]
SOPHIE: I know it's not easy to say for sure. I would like to be fully prepared for 
        the future though. Is there really no way to tell what my tests mean for me?
User:   Absolutely. I think that what I have to say might be a little difficult to
        hear, but it seems like you want to know. Based on the test, it looks like 
        the cancer really has not responded to the treatment, and that there may not
        be any further treatment available to shrink the tumor.
SOPHIE: I am sorry, I'm having trouble understanding. Could you try saying it again, 
        a bit more clearly?
User:   Sure. I was just saying that, as you suspected from the scan, it looks like
        the radiation treatment was not entirely successful, and that the tumor size
        has not shrunk.
SOPHIE: I'm not surprised to hear the radiation did not help, as my pain has gotten
        worse in these past couple of weeks. But what does this mean for my future?
        How long do I have left?
...
\end{lstlisting}
        \caption{A ``lemon-picked'' excerpt from the SOPHIE pilot study in which the system fails to extract a gist-clause from the user's input, causing it to employ a clarification request fallback so that the user rephrases their utterance in a way that the system understands.}
        \label{fig:sophie_bad_transcript1}
     \end{subfigure}
     \begin{subfigure}{\textwidth}
        \small
\begin{lstlisting}[language=DialogueExamples,frame=single,mathescape=true]
SOPHIE: I'm still not really sure how I should break this news to my family. I don't
        want them to be worried .
User:   Would it help to have some more information about what the scan showed?
SOPHIE: Sorry, I didn't catch that. Could you repeat that one more time using a
        different phrasing?
User:   Would you like to talk about the results of the ct scan?
SOPHIE: I'm not hearing you very well. Would you mind speaking more clearly?
User:   Would you like to talk about the results of the ct scan?
SOPHIE: Sorry, I still didn't quite understand. I have some more questions, maybe
        we can come back to it.
...
\end{lstlisting}
        \caption{A ``lemon-picked'' excerpt from the SOPHIE pilot study in which the system generates multiple clarification requests, causing the user to simply repeat the utterance that lead to a gist-clause failure and significantly disrupting the conversation.}
        \label{fig:sophie_bad_transcript2}
     \end{subfigure}
     \begin{subfigure}{\textwidth}
        \small
\begin{lstlisting}[language=DialogueExamples,frame=single,mathescape=true]
SOPHIE: I'm just scared about it getting worse, and not being able to be there for 
        my family.
User:   I hear that family is very important to you.
SOPHIE: I suppose these things can be hard to predict.
...
\end{lstlisting}
        \caption{A ``lemon-picked'' excerpt from the SOPHIE pilot study in which the system generates a default generic reaction due to a failure to extract a gist-clause. However, the particular default reaction given in this case is an inappropriate response to the empathetic response from the user.}
        \label{fig:sophie_bad_transcript3}
     \end{subfigure}
     \caption{}
     \label{fig:sophie_bad_transcripts}
\end{figure*}




\end{document}